\documentclass[runningheads]{llncs}

\usepackage{eccv}

\usepackage{eccvabbrv}

\usepackage{graphicx}
\usepackage{booktabs}
\usepackage{wrapfig}

\usepackage[accsupp]{axessibility}  

\usepackage[pagebackref,breaklinks,colorlinks,citecolor=eccvblue]{hyperref}

\usepackage{orcidlink}

\begin{document}

\title{POLO - Point-based, multi-class animal detection} 


\author{Giacomo May~\inst{1} \and
Emanuele Dalsasso\inst{1}\orcidlink{0000-0001-7170-9015} \and\\
Benjamin Kellenberger\inst{2}\orcidlink{0000-0002-2902-2014} \and Devis Tuia\inst{1}\orcidlink{0000-0003-0374-2459}}

\authorrunning{G. May et al.}

\institute{EPFL, Switzerland
\email{(first.last@epfl.ch)} \and
University College London, U.K. (\email{b.kellenberger@ucl.ac.uk})
}

\maketitle

\begin{abstract}
Automated wildlife surveys based on drone imagery and object detection technology are a powerful and increasingly popular tool in conservation biology. Most detectors require training images with annotated bounding boxes, which are tedious, expensive, and not always unambiguous to create. To reduce the annotation load associated with this practice, we develop POLO, a multi-class object detection model that can be trained entirely on point labels. POLO is based on simple, yet effective modifications to the YOLOv8 architecture, including alterations to the prediction process, training losses, and post-processing. We test POLO on drone recordings of waterfowl containing up to multiple thousands of individual birds in one image and compare it to a regular YOLOv8. Our experiments show that at the same annotation cost, POLO achieves improved accuracy in counting animals in aerial imagery.  
  \keywords{Animal detection \and Wildlife censuses \and Annotation cost}
\end{abstract}

\section{Introduction}
\label{sec:intro}
Frequent animal censuses are a key requirement for successful conservation management and become particularly important when dealing with endangered species. Wildlife in vast and open landscapes can be surveyed efficiently using aerial imagery recorded either from airplanes or unmanned aerial vehicles (UAVs), with the latter being used more and more thanks to the reduced operating costs and safety risks \cite{kellenberger2018detecting, linchant2015unmanned}. Given the large amounts of data collected during these flights, machine learning approaches are often employed to count animals in the images, allowing biologists to estimate the development of populations. For this purpose, convolutional neural networks (CNNs) are among the most popular techniques \cite{kellenberger2017fast, dujon2021machine, delplanque2022multispecies, eikelboom2019improving}.\\
While CNNs hold the promise of high detection accuracy, this potential is limited by the volume of labeled data available for model training \cite{alzubaidi2021review}. Since the creation of labeled data implies manual annotation, often to be provided in the form of bounding boxes, accurate counting of animals from aerial imagery comes at a significant annotation cost, which limits the scalability of deep learning-based conservation efforts. To reduce these costs, bounding boxes can be automatically created from point annotations, which can be obtained with higher speed and are thus much cheaper to produce \cite{mullen2019comparing, ge2023point}. This approach simply consists of generating square boxes around the point annotations and using these pseudo-labels to train conventional object detection architectures. However, animals in aerial images are regularly of very small size (\ie, a few pixels in length), partially occluded, difficult to separate if standing close together, or distorted due to perspective and motion blur. All of these factors contribute to animal boundaries being difficult to demarcate, hence significantly affecting the quality of automatically generated boxes.\\
In the present study, we avoid these issues without raising the annotation cost by developing an object detection framework that can be trained entirely on point labels. We approach this task by modifying the YOLOv8 algorithm \cite{Jocher_Ultralytics_YOLO_2023}, one of the most advanced and widely used bounding box detection models. The proposed architecture is called POLO (\textbf{P}oint-Y\textbf{OLO}). We compare two different loss functions to train it, and introduce a new metric to perform non-maximum-suppression (NMS) on point predictions.\\
We test POLO on a challenging dataset, which involves counting five species of waterfowl in drone images of the Izembek lagoon in Alaska. Moreover, we compare POLO's performance to that of a regular YOLOv8 trained on pseudo-labels, and find that while the latter approach yields high counting accuracy, results improve when using POLO.

\section{Related Work}
\subsection{Point-based object detection and counting}
Point labels represent a popular annotation type in the field of crowd counting, where they are used to train density estimation models that take images as input and output a heat map encoding the estimated density of people in every pixel \cite{boominathan2016crowdnet, zhao2016crossing, li2018csrnet} -- an approach motivated by the observation that conventional detection algorithms struggle with scale variation (\ie, people in the background appearing smaller than people close to the camera) and partial occlusion of individuals by other individuals.\\
In object detection on the other hand, the point format has been used as part of mixed labeled sets, where a small set of bounding box annotations is complemented by  a larger set of point labels to train detection models to ouptut bounding box predictions \cite{ge2023point, zhang2022group, chen2021points}. \\
\\
Our solution differs from the above methods, since we output point detections (rather than density maps or bounding boxes) and do not use any bounding box annotations for training.  This makes for a simpler learning task than generating bounding box outputs from point labels, and suffices for our purpose of counting animals in drone imagery. 
The present work is most similar to the efforts of Song \etal \cite{song2021rethinking}, who also develop a detection model that trains on and outputs point labels. However, their architecture cannot distinguish between different classes, making it unsuitable for animal censuses involving multiple species.

\subsection{The YOLO algorithm} \label{sec: YOLO}
To identify objects, YOLOv8 employs a one-stage detection strategy that divides images into a predefined number of grid cells. Each grid cell is then processed on two separate branches, one predicting bounding boxes that enclose objects lying within the cell, and the other predicting the class of said objects. Importantly, YOLOv8 does not directly regress bounding box coordinates, but samples a probability distribution to obtain the most likely number of pixels by which the four bounding box edges are offset from the center of grid cells.\\
The training objective of YOLOv8 consists of three loss terms: a \textit{binary cross-entropy} loss ($\mathcal{L}_{BCE}$) for the class predictions, an \textit{Intersection-over-Union} loss ($\mathcal{L}_{IoU}$) to learn the geometric prediction of bounding boxes, and a \textit{Distribution Focal Loss} ($\mathcal{L}_{DFL}$) \cite{li2022generalized} to optimize the probability distribution used for predicting bounding box offsets. The overall loss value is then computed as a weighted sum of these three components, where the class, IoU, and DFL loss are scaled by factors 0.5, 7.5, and 1.5, respectively.

\section{Method} \label{sec: methods}
To achieve compatibility with point labels, we apply the following modifications to the YOLOv8 architecture:
\begin{enumerate}
    \item \textbf{Output dimensions:} By default, the number of output channels in the final convolutional layer of YOLOv8 is $4 \cdot 16$. These channels encode probabilities of the different bounding box edges being offset by $[0, 1, ..., 15]$ pixels from the center of a grid cell.\\
    We change the architecture to predict the x- and y- coordinates of the center points of objects that lie within grid cells, using only two channels:
    \begin{align}
        \hat{p}_x &= \sigma(a_1) \cdot 2 - 0.5 + c_{x}\\
        \hat{p}_y &= \sigma(a_2) \cdot 2 - 0.5 + c_{y}
    \end{align}
    where $\hat{p}_x$ and $\hat{p}_y$ are the predicted coordinates, $a_1$ and $a_2$ are the activation values of a grid cell in the first and second output channel, $\sigma(\cdot)$ is the sigmoid function, and $c_x$ and $c_y$ are the coordinates of the grid cell's top left corner. This way, predicted values can range between -0.5 and 1.5, allowing the model to locate objects that do not fit entirely into one cell. We borrow this approach from YOLOv8's predecessor, YOLOv5 \cite{Jocher_Ultralytics_YOLO_2023}, where it is used to regress the center point of bounding boxes.\\
    \item \textbf{Loss function:} Neither the IoU-, nor the DFL-loss term can be applied to our point model. We consider the following alternatives:
    \begin{enumerate}
        \item Average Hausdorff-Distance \cite{ribera2019locating}: 
            \begin{equation}
                \mathcal{L}_{AH}(\hat{P}, P) = \frac{1}{|P|}\sum_{i=1}^{|P|}\texttt{min}_{\hat{\mathbf{p}} \in \hat{P}} d(\hat{\mathbf{p}}, \mathbf{p}_{i}) + \frac{1}{|\hat{P}|}\sum_{j=1}^{|\hat{P}|}\texttt{min}_{\mathbf{p} \in P} d(\hat{\mathbf{p}}_{j}, \mathbf{p}) 
            \end{equation}
            where $\hat{P}$ is the set of predicted points, $P$ is the set of ground truth locations, $|\cdot|$ denotes the cardinality, and $d(\cdot, \cdot)$ is the Euclidean distance.
        \item Mean Squared Error (MSE) \cite{song2021rethinking}:
            \begin{equation}
                \mathcal{L}_{MSE} = \frac{1}{|P|} \sum_{i=1}^{|P|} ||\mathbf{p}_{i} -\hat{\mathbf{p}}_{i}||_{2}^{2}
            \end{equation}
            Again, $P$ denotes the set of ground truth locations and $\hat{\mathbf{p}}_{i}$ is the prediction corresponding to ground truth $\mathbf{p}_{i}$.\\
    \end{enumerate}
    \item \textbf{Postprocessing:} To remove redundant detections, we implement a variation of the traditional NMS algorithm. Specifically, we define the \textit{Distance-over-Radius} (DoR) metric, and use it to replace the IoU. Here, the DoR is computed by dividing the distance of a predicted point to its ground truth label by a radius value $r_{c}$ specified by the user for each object/animal class in the dataset:

    \begin{equation}\label{eq: DoR}
        DoR = \frac{d(\hat{\mathbf{p}}, \mathbf{p})}{r_{c}}
    \end{equation}

    with $d(\hat{\mathbf{p}}, \mathbf{p})$ denoting  the Euclidean distance between a predicted point and the ground truth location. 
    During NMS, low-confidence detections are removed if their DoR to higher-confidence detections falls below a specified threshold. 
    
\end{enumerate}

\section{Experiments}
\subsection{Experimental Setup}
\subsubsection{Dataset}
We evaluate and test all models in this study on a publicly available set of drone images \cite{weiser2022counts} from the Izembek lagoon in Alaska hosted on the \href{https://lila.science}{LILA BC data repository}. The dataset includes 9,2672 images of waterfowl of size $8688\times5792$ pixels, and a total of 521,270 pseudo-boxes for the classes ``Brant goose'' (424,790 boxes), ``Canada goose'' (47,561 boxes), ``Gull'' (5,631 boxes), ``Emperor goose'' (2,013 boxes), and ``Other'' (5,631 boxes). The data was divided into a training (80\%), validation (5\%), and test set (15\%). Images were split into $640\times640$ patches with 10\% overlap, to match the input size expected by YOLOv8 and POLO. 95\% of patches that did not contain animals were discarded. The remaining 5\% were used as negative samples to reduce false positive detections. Bounding boxes that span multiple patches were clipped to the patch limits if at least 15\% of the area of the box lied within the patch in question.

\subsubsection{Implementation Details} \label{sec: implementation_details}
Throughout experiments, we employ a batch size of 32 and set the training duration to 300 epochs, while activating YOLO's early stopping mechanism with a patience value of 50. Counting accuracy was assessed by applying the trained models to our test set. There too, we split images into 640 $\times$ 640 patches with 10\% overlap, but map the patch-level predictions back to image-level after inference to obtain a global animal count for the entire image. An additional round of NMS is applied after mapping patch predictions to the image-level in order to remove redundancy in the patch overlap regions.

\subsection{Loss functions comparison} \label{sec: loss_comp}
In a first step we compare the counting accuracy achieved with POLO models trained on the Hausdorff and MSE loss, respectively. We set a DoR threshold of 0.3 for post-processing during this initial comparison, and use the YOLOv8 loss balancing scheme mentioned in \cref{sec: YOLO}; \ie, we assign the Hausdorff-/MSE-loss a weight of 7.5, and scale the classification loss by 0.5. We use a 40 pixel radius for all classes, except for the Gull category, where we set the radius to 30 pixels. These values were obtained by manually measuring the length of animals in the training images, and adding a buffer of 10-15 pixel to account for variations in ground sampling distance between images, and in the appearance of birds. For example, when in flight, animals will occupy more pixels compared to when they are resting on the water surface due to being closer to the drone and their wings being spread.
\cref{tab:counts} displays the mean absolute error (MAE) per image obtained when training POLO with the Hausdorff/MSE loss function. 

\begin{table}[!h]
    \centering
    \begin{tabular}{ l || c | c | c | c | c }
        & Brant Goose & Other & Gull & Canada Goose & Emperor Goose\\  
        \hline
        \hline
        POLO MSE & \textbf{5.91} & \textbf{4.57} & 0.93 & 2.26 & 0.25 \\
        POLO Hausdorff & 6.57 & 4.64 & \textbf{0.91} & \textbf{2.19} & \textbf{0.22}\\
    \end{tabular}
    \caption{MAE scores obtained with the MSE/Hausdorff loss function. \vspace{-2em}}
    \label{tab:counts}
\end{table}

As can be seen, the Hausdorff model beats the MSE model in three out of five categories. However, for the most abundant species of the dataset (Brant goose), the MSE loss reduces the MAE considerably. We hence decided to use the latter for all subsequent experiments. 

\subsection{Loss balancing} \label{sec: loss_balance}

\begin{table}
\centering
    \begin{tabular}{ c || c | c | c | c | c }
        \textbf{$\alpha$} & Brant Goose & Other & Gull & Canada Goose & Emperor Goose \\
        \hline
        \hline
        1 & \textcolor{teal}{5.3} & \textcolor{orange}{4.17} & \textcolor{teal}{0.73} & \textcolor{teal}{1.89} & \textcolor{orange}{0.22}\\
        \hline
        2 & 5.58 & \textcolor{teal}{4.11} & 0.86 & \textcolor{orange}{1.94} & 0.22 \\
        \hline
        3 & 5.48 & 4.2 & \textcolor{orange}{0.74} & 2.03 & 0.2 \\
        \hline
        4 & 5.51 & 4.23 & \textcolor{green}{\textbf{0.69}} & \textcolor{green}{\textbf{1.81}} & \textcolor{teal}{0.2} \\
        \hline
        5 & 5.34 & \textcolor{green}{\textbf{4.07}} & 0.86 & 1.95 & 0.22 \\
        \hline
        6 & 5.63 & 4.31 & 0.78 & 2.02 & 0.21 \\
        \hline
        7 & \textcolor{green}{\textbf{5.25}} & 4.77 & 0.83 & 2.05 & \textcolor{green}{\textbf{0.19}} \\
        \hline
        8 & \textcolor{orange}{5.31} & 4.28 & 0.81 & 2.02 & 0.22 \\
        \hline
        9 & 5.72 & 4.62 & 0.79 & 1.97 & 0.23
    \end{tabular}
    \caption{MAE scores for varying loss weights. The \textcolor{green}{\textbf{first}}, \textcolor{teal}{second}, and \textcolor{orange}{third} best scores are colored in \textcolor{green}{green}, \textcolor{teal}{blue}, and \textcolor{orange}{orange}, respectively. \vspace{-2em}}
    \label{tab: loss weights}
\end{table}

To optimize the balance between the MSE and classification loss terms, we introduce a hyperparameter $\alpha \in [1, 9]$, based on the value range used for loss scaling in YOLOv8. We then train a total of nine different models using the DoR threshold and radii specified in \cref{sec: loss_comp}, where the loss is composed as follows: 
\begin{equation}
    \mathcal{L} = \alpha \cdot \mathcal{L}_{MSE} + (10 - \alpha) \cdot \mathcal{L}_{BCE}
\end{equation}
\cref{tab: loss weights} contains the MAE scores for different values of $\alpha$. We find the model trained with $\alpha = 1$  to perform consistently well across classes: while it is never the most accurate, it yields the second- or third-lowest MAE score in every category. Consequently, we choose the model trained with $\alpha=1$ for the remaining experiments.

\subsection{Radius and DoR Threshold} \label{sec: radii_dor}
\begin{wrapfigure}{R}{0.7\textwidth}\vspace{-2em} \centering \scriptsize 

    \includegraphics[width=0.6\textwidth]{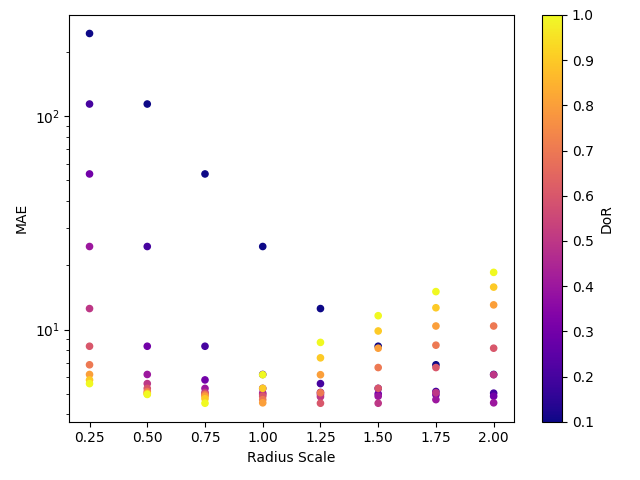}
    \caption{MAE scores achieved for the Brant goose class depending on radius and DoR value. \vspace{-3em}}
    \label{fig: radii_dor}
\end{wrapfigure}
We finally perform a full grid search over combinations of radius values and DoR thresholds. We probe DoR thresholds between [0.1, 1] and multiply the radii defined in \cref{sec: implementation_details} by scaling factors in the range of [0.25, 2]. \cref{fig: radii_dor} visualizes the effect different radii and DoR thresholds have on the MAE achieved for the Brant goose class. As can be seen, two strategies maximize counting accuracy: using intermediate values for both, radius and DoR threshold, or pairing opposite extremes; \ie, combining large radii with low DoR tresholds and vice versa. It should be noted though that the MAE is affected less strongly when large radii are used with high DoR thresholds, compared to the combination of small radii and low DoR values. We observe similar behavior in the remaining classes, where a scaling factor of 1.25 and a DoR-threshold of 0.6 achieve top MAE scores across categories.

\section{Results} \label{sec: results}
The animal counts and MAE values obtained with YOLOv8 and POLO are showcased in \cref{tab: counts_final}. Based on the above results, we use a POLO model trained with the MSE loss function scaled by factor 1 in this experiment. We multiply the radii mentioned in \cref{sec: loss_comp} by factor 1.25, and set the DoR threshold to 0.6. 
\begin{table}[!h]
    \scriptsize
    \centering
    \begin{tabular}{ l || c | c | c | c | c | c | c | c | c | c}
    & \multicolumn{2}{|c|}{Brant Goose} & \multicolumn{2}{|c}{Other} & \multicolumn{2}{|c}{Gull} & \multicolumn{2}{|c}{Canada Goose} & \multicolumn{2}{|c}{Emperor Goose}\\
        \cline{2-11}
         & Count &  MAE  & Count &  MAE & Count &  MAE & Count &  MAE & Count &  MAE\\
        \hline
        \hline
        YOLOv8 & 68,732 & 5.86 & 6,256 & \textbf{3.78} & 1,251 & 1.05 & 7,113 & 1.98 & 213  & 0.26 \\
        POLO & 67,501 & \textbf{4.51} & 6,436 & 4.08 & 961 & \textbf{0.69} & 6,923 & \textbf{1.87} & 149 & \textbf{0.22} \\
        \hline
        \textbf{Ground Truth} & \multicolumn{2}{|c|}{64,764} &  \multicolumn{2}{|c|}{4,910} &  \multicolumn{2}{|c|}{584} &  \multicolumn{2}{|c|}{7,233} & \multicolumn{2}{|c}{225}\\
    \end{tabular}
    \caption{Counting accuracy of YOLOv8 and POLO.\vspace{-3em}}
    \label{tab: counts_final}
\end{table}

\begin{figure}[!htb]
      \centering
	   \begin{subfigure}{0.24\linewidth}
		\includegraphics[width=\linewidth, height=2cm]{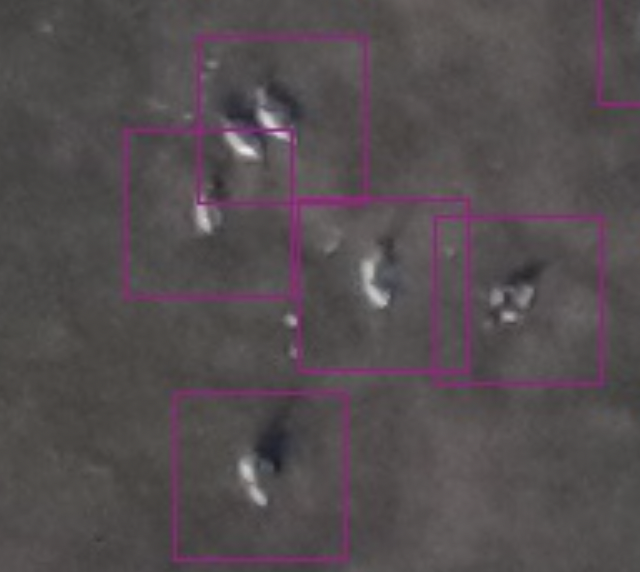}
		\label{fig:TP_YOLO}
	   \end{subfigure}
	   \begin{subfigure}{0.24\linewidth}
		\includegraphics[width=\linewidth, height=2cm]{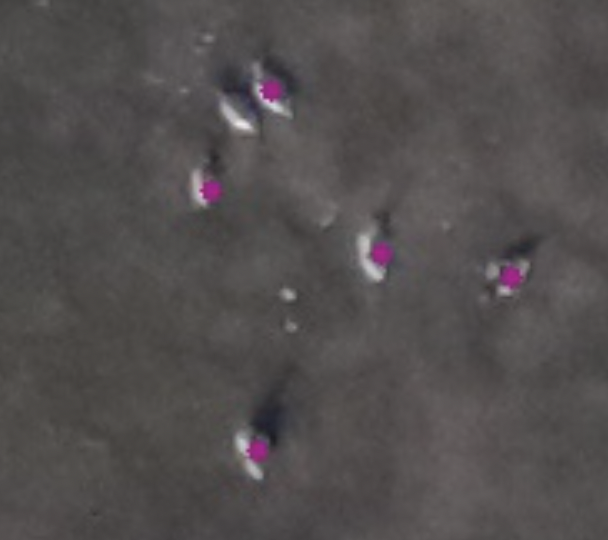}
		\label{fig:TP_POLO}
	    \end{subfigure}
	     \begin{subfigure}{0.24\linewidth}
		 \includegraphics[width=\linewidth, height=2cm]{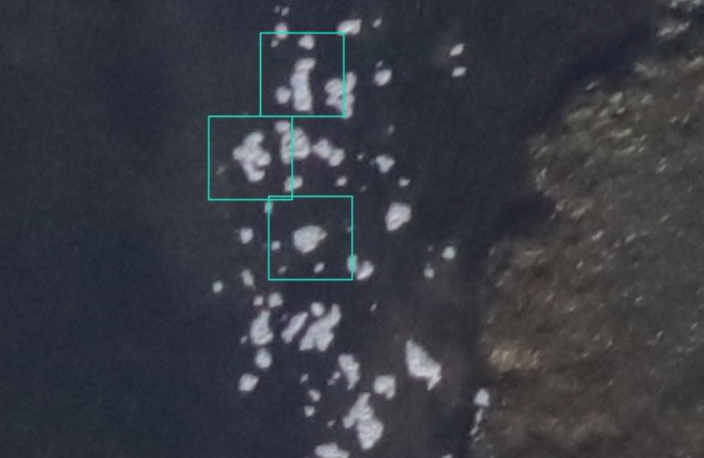}
		 \label{fig:FP_YOLO}
	      \end{subfigure}
	       \begin{subfigure}{0.24\linewidth}
		  \includegraphics[width=\linewidth, height=2cm]{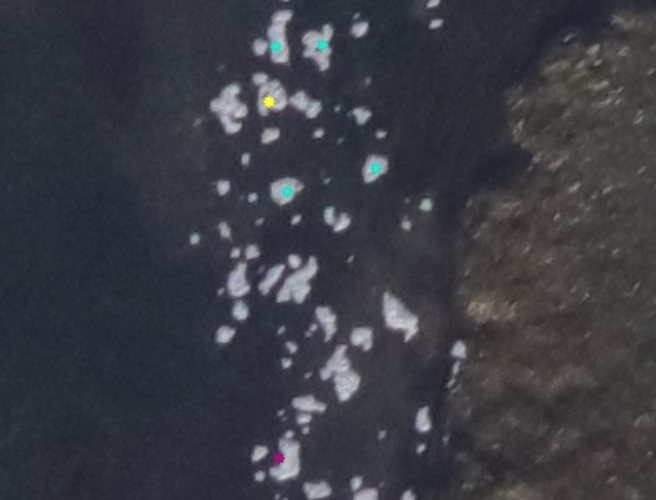}
		  \label{fig:FP_POLO}
	       \end{subfigure}
	\caption{True- (columns 1 \& 2) and false-positive (columns 3 \& 4) detections obtained with YOLOv8 and POLO (magenta = Brant Goose, turquoise = Other, yellow = Gull).\vspace{-2em}}
	\label{fig: vis_examples}
\end{figure}

Overall, both models yield reasonable counts, but overestimate the abundance of the classes ``Brant Goose'', ``Other'', and ``Gull'' is overestimated , while under-detecting ``Canada Goose'' and ``Emperor Goose''. Importantly, POLO achieves a lower MAE in four out of five categories.\\
Qualitatively, both architectures struggle to separate animals in close proximity of each other, and tend to mistake bright water structures for birds. Differences lie mostly in the classes assigned to these false positive predictions (\cf \cref{fig: vis_examples}). 

\section{Conclusion}
Our results show that YOLOv8 achieves surprisingly accurate animal counts when trained on pseudo-labels. However, we manage to improve counting accuracy with POLO, and obtain consistently lower MAE scores across categories. As this leads to more conservative estimates compared to YOLOv8, POLO appears less prone to false positive detections. 
The modifications proposed in this work hence represent an improved way to minimize the annotation cost of conventional object detection. Subsequent efforts will be directed towards assessing the difference between POLO and YOLOv8 when hand-crafted bounding boxes are used for training the latter. We further plan to study the behavior of POLO under different data acquisition scenarios, such as flight altitude, camera angle, etc. 

\section*{Funding Information}
This research has been carried out as part of the project WildDrone, funded by the European Union's Horizon Europe Research and Innovation Program under the Marie Skłodowska-Curie Grant Agreement No. 101071224, the EPSRC funded Autonomous Drones for Nature Conservation Missions grant (EP/X029077/1), and the Swiss State Secretariat for Education, Research and lnnovation (SERI) under contract number 22.00280.

%
%
\bibliographystyle{splncs04}
\bibliography{main}
\end{document}